\documentclass[letterpaper]{article} 
\usepackage[preprint]{aaai2027} 
\usepackage[hyphens]{url} 
\usepackage{graphicx} 
\usepackage{natbib} 
\usepackage{caption} 
\usepackage{amsmath,amssymb}
\usepackage{booktabs}
\usepackage{array}
\usepackage{tikz}
\usetikzlibrary{arrows.meta,positioning,shapes.geometric,fit}
\usepackage{xcolor}

\definecolor{AIRIpaper}{HTML}{F2EEE9}
\definecolor{AIRIteal}{HTML}{2FBEAD}
\definecolor{AIRInoonteal}{HTML}{3DEBCB}
\definecolor{AIRIgreen}{HTML}{005A60}
\definecolor{AIRIdarkblue}{HTML}{161C27}
\definecolor{AIRIblue}{HTML}{2DAAF0}

\newcolumntype{P}[1]{>{\raggedright\arraybackslash}p{#1}}
\newcommand{\widefig}[1]{\includegraphics[width=\linewidth]{#1}}
\newcommand{\midwidefig}[1]{\includegraphics[width=\linewidth]{#1}}
\graphicspath{{figures/}}

\title{Call-Efficient ICD-10 Code Prediction via Supervised Candidate Generation and Selective LLM Invocation}
\author{
Ilia Kopanichuk\textsuperscript{\rm 1,2}, Daniil Kartashev\textsuperscript{\rm 1}, Yuri Bolkonsky\textsuperscript{\rm 1}, Petr Anokhin\textsuperscript{\rm 1,5},\\
Vladimir Makharev\textsuperscript{\rm 1,4}, Iaroslav Bespalov\textsuperscript{\rm 1}, Dmitry V. Dylov\textsuperscript{\rm 1,3}
}
\affiliations{
\textsuperscript{\rm 1}AIRI, Moscow, Russia\\
\textsuperscript{\rm 2}Moscow Institute of Physics and Technology, Dolgoprudny, Russia\\
\textsuperscript{\rm 3}Skolkovo Institute of Science and Technology, Moscow, Russia\\
\textsuperscript{\rm 4}Innopolis University, Innopolis, Russia\\
\textsuperscript{\rm 5}Lomonosov Moscow State University, Moscow, Russia\\
\texttt{kopanichuk@airi.net}
}

\begin{document}
\maketitle

\begin{abstract}
Automatic assignment of ICD-10 codes from short clinical diagnosis text is a practical medical text classification and normalization task. The task is difficult because diagnosis phrases are short and noisy, the ICD-10 label space is hierarchical, and the class distribution is strongly imbalanced. This paper presents a call-efficient ICD-10 code prediction pipeline that combines transformer embeddings, tree-based supervised candidate generation, candidate-level reranking, and a gated LLM assistance. The evaluation uses a leakage-safe validation protocol: model and threshold selection are based on out-of-fold cross-validation predictions, while final reporting is performed on a locked test set with no shared source-file groups. The supervised conditional parent-code candidate generator reaches 74.32\% Top-1 accuracy on the locked test set and 93.05\% Top-10 candidate coverage. A direct zero-shot GPT-5.6-Sol baseline reaches 81.01\% Top-1 accuracy, but requires an invocation of this large model for every diagnosis. The final deployable system, Pre-LLM fusion, reaches 81.24\% OOF-CV accuracy and 83.19\% locked-test accuracy while calling the LLM for only 31.97\% of cases. This improves the locked-test error rate from 25.68\% for the supervised parent-code baseline to 16.81\%, corresponding to a 34.5\% relative error reduction. The results show that the most effective role of the LLM is controlled reranking on uncertain examples rather than unrestricted coding or full-dataset invocation.

\end{abstract}

\section{Introduction}
Automatic assignment of ICD-10 codes from clinical narratives can reduce the burden of manual coding and improve the scalability of downstream analytics~\cite{perotte2014diagnosis,mullenbach2018explainable}. Diagnosis codes are used for reporting, reimbursement, cohort construction, epidemiological analysis, quality control, and downstream decision-support tasks. A reliable coding system should therefore balance predictive accuracy with inference requirements while remaining easy to audit.

The task is difficult because it combines language understanding with a structured and imbalanced target space. A diagnosis phrase can be short, abbreviated, partially misspelled, or written with local terminology. ICD-10 is hierarchical: the first letter defines a broad diagnostic group, while subsequent characters define increasingly specific conditions. A model may predict the correct broad letter but fail to select the exact parent code, or it may confuse the broad group entirely.

A second practical challenge is the use of large language models. LLMs can improve candidate selection, but calling them for every diagnosis may be expensive and unnecessary. The final system studied here is therefore not an unrestricted LLM coder. Instead, a supervised model first produces a high-recall top-10 parent-code candidate list. The LLM is used only when a learned pre-LLM gate predicts that it is likely to help, and a lightweight candidate ranker provides an additional no-call alternative. The final fusion selector combines these decisions.

The main contribution of this paper is an end-to-end evaluation of gated LLM assistance on top of a supervised candidate-generation pipeline. We report both OOF-CV and locked-test results and evaluate predictive accuracy jointly with the LLM invocation rate. The final Pre-LLM fusion approach reaches 83.19\% locked-test Top-1 accuracy with 31.97\% LLM calls.

\begin{table*}[!t]
\centering
\small
\begin{tabular}{@{}P{0.24\linewidth}P{0.38\linewidth}P{0.30\linewidth}@{}}
\toprule
Work or system & Published setting and reported result & Relation to the present study \\
\midrule
RuCCoD BERT pipeline~\cite{ruccod2025} & Russian NER plus entity-level ICD-10 assignment; F1 = 0.525 and accuracy = 0.356. & The closest Russian academic benchmark; our work studies direct ICD-10 letter and parent-code prediction from diagnosis text. \\
Phi-3.5-mini + ICD dictionary + RuCCoD~\cite{ruccod2025} & LLM tuning with retrieval-augmented ICD assignment; F1 = 0.480 and accuracy = 0.316. & Represents an LLM/RAG alternative; our final method uses a supervised candidate generator and calls the LLM selectively. \\
LLaMA3-8B-Instruct + ICD dictionary + RuCCoD~\cite{ruccod2025} & Zero-shot retrieval-augmented ICD assignment; F1 = 0.458 and accuracy = 0.297. & Shows that retrieval helps LLM baselines, but supervised candidate generation remains important for Russian ICD linking. \\
RuBioBERT / RuBioRoBERTa ~\cite{yalunin2022rubio} & Russian biomedical encoders; RuCCoD reports F1 = 0.77 for diagnosis-entity extraction. & Useful for entity recognition and representation learning; our experiments use transformer embeddings and tree-based classifiers. \\
Russian ICD-10 chapter classifier and per-letter subgroup model series~\cite{hfchapter,hfsubgroups} & Public short-text classifiers; the chapter model reports top-1 accuracy = 0.7121 and top-3 accuracy = 0.8724. & These models predict coarse chapters or letter-specific subgroups, while our analysis focuses on parent-code prediction and selective LLM use. \\
\bottomrule
\end{tabular}
\caption{Closest Russian-language ICD-10 coding analogues and the position of the present study. The results are not ranked against the present model because the datasets and label spaces differ.}
\label{tab:russian_analogues}
\end{table*}

\section{Related work}
Automatic medical coding has been studied in several related formulations. The most common English-language formulation is multi-label document classification: a model receives a long clinical note, such as a discharge summary, and predicts all ICD codes assigned to the document. Classic neural systems for this setting include convolutional and label-attention models such as CAML, MultiResCNN, and LAAT~\cite{mullenbach2018explainable,li2020multirescnn,vu2020laat}. More recent systems use pretrained language models and explicit code knowledge. PLM-ICD applies pretrained encoders to long clinical documents, KEPTLongformer injects domain knowledge during pretraining, and GKI-ICD incorporates descriptions, synonyms, and hierarchy information for ICD coding~\cite{huang2022plmicd,yang2022kept,gki2025}.

A second formulation is closer to practical short-text diagnosis normalization. Instead of predicting many codes from a long document, the system receives a short diagnosis phrase and returns the most appropriate ICD code or a small list of candidates. This setting is related to biomedical concept normalization, medical entity linking, and ICD retrieval. It is especially relevant for Russian clinical text, where diagnosis phrases may contain abbreviations, spelling variation, informal wording, and incomplete specificity. Retrieval-based pipelines encode both the input diagnosis and dictionary entries, then select candidates by nearest-neighbor search or reranking~\cite{sapbert2021,coder2022,bergamot2024}.

For Russian ICD coding, the closest academic benchmark is RuCCoD. RuCCoD formulates the task as a pipeline of nested entity recognition, entity linking, and record-level code aggregation. The best reported BERT-based entity-level ICD assignment result reaches F1 = 0.525 and accuracy = 0.356, while the strongest LLM-based variant reported in the same comparison remains lower, with F1 = 0.480 and accuracy = 0.316~\cite{ruccod2025}. Public Russian Hugging Face checkpoints also include a chapter classifier and a family of per-letter ICD-10 subgroup classifiers, but they solve coarser or narrower tasks than full ICD-10 linking~\cite{hfchapter,hfsubgroups}. The closest Russian-language analogues are summarized in Table~\ref{tab:russian_analogues}.

\section{Data and validation protocol}

The input data consist of short clinical diagnosis formulations in Russian. Each sample is associated with a full ICD-10 code, a three-character parent code, and an ICD-10 letter. The study focuses on two nested prediction tasks. The first task is letter-level prediction, which evaluates whether the model assigns the correct ICD-10 letter. The second task is parent-code prediction, which is more specific and corresponds to selecting a code such as \texttt{J02}, \texttt{L20}, or \texttt{M54}. This decomposition is useful because the letter stage determines the broad diagnostic chapter, while the parent stage provides the clinically actionable category.

The final validation split contains 8,552 training rows and 1,511 locked-test rows. It was selected without using model accuracy: the split search minimized distribution drift and leakage-related criteria. The selected split has zero shared \texttt{source\_file} groups between train and test, letter Jensen-Shannon divergence of 0.0030, parent-code divergence of 0.0867, and code divergence of 0.1951. 

Model and threshold selection are based on five out-of-fold predictions on the training split. The locked test set is used only for final reporting. This separation is important because both the LLM gate and the fusion selector can otherwise overfit to whether the LLM helps on a particular row.

\begin{table}[htbp]
\centering
\small
\begin{tabular}{@{}lcc@{}}
\toprule
 & Train & Locked test \\
\midrule
Size of sample & 8,552 & 1,511 \\
Number of unique parent codes & 637 & 330 \\
\bottomrule
\end{tabular}
\caption{Train-test dataset statistics for the final split. Parent codes correspond to the three-character ICD-10 code prefixes.}
\label{tab:stats}
\end{table}

\begin{figure}[htbp]
  \centering
  \midwidefig{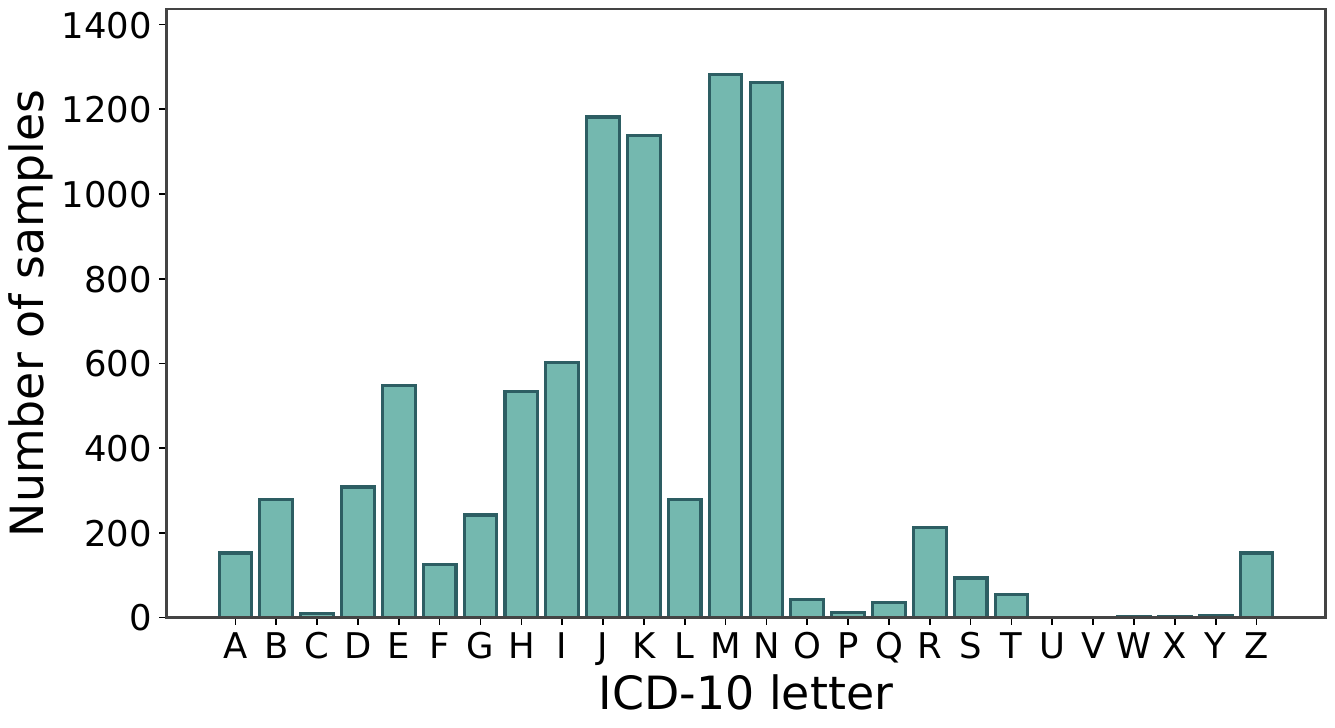}
  \caption{Training-set distribution by ICD-10 letter. The frequencies are strongly imbalanced, with letters such as J, K, M, and N dominating the dataset.}
  \label{fig:training_distribution}
\end{figure}

The class distribution remains strongly imbalanced. Figure~\ref{fig:training_distribution} shows that a small number of letters dominate the training data, while many letters have much smaller support. This motivates macro diagnostics, per-class metrics, and separate rare-versus-frequent analysis.

\section{Methodology}
\subsection{Supervised prediction pipeline}
The semantic matching component of our pipeline is adapted from~\cite{kopanichuk2025evaluate}, who introduced a supervised match function for determining whether two free-form diagnostic expressions refer to the same clinical condition. Their original meta-model combines direct LLM judgments, ICD-10 retrieval, embedding similarities, and lexical features, and was subsequently distilled into a ModernBERT sequence-pair classifier for lower-cost inference. Given two textual diagnosis descriptions \(d_i\) and \(d_j\), we use the continuous output of this distilled matcher as their pairwise semantic score \(s(d_i,d_j)\), which is used to construct hierarchical ICD-10 representations for the downstream letter and parent-code classifiers.

The supervised predictor consists of text preprocessing, embedding extraction with ModernBERT, letter-level classification with Random Forest, and category classification with HistGradientBoosting. The category stage is conditional on the ICD-10 letter, which means Random Forest classifier predicts the letter, then HistGradientBoosting classifier takes predicted letters as inputs and predicts the category. This allows the model to select among categories within the relevant letter. In the hybrid setting, the supervised candidate list is passed to an LLM reranker that reorders the candidates but does not generate codes outside the candidate set.

\begin{figure*}[htbp]
\centering
\begin{tikzpicture}[
    input/.style={rectangle, rounded corners, draw=AIRIgreen, fill=AIRIpaper, text=AIRIdarkblue, align=center, minimum width=2.0cm, minimum height=0.9cm, font=\scriptsize},
    supervised/.style={rectangle, rounded corners, draw=AIRIgreen, fill=AIRIpaper, text=AIRIdarkblue, align=center, minimum width=2.0cm, minimum height=0.9cm, font=\scriptsize},
    candidates/.style={rectangle, rounded corners, draw=AIRIgreen, fill=AIRIpaper, text=AIRIdarkblue, align=center, minimum width=2.3cm, minimum height=0.9cm, font=\scriptsize},
    nocall/.style={rectangle, rounded corners, draw=AIRIgreen, fill=AIRIpaper, text=AIRIdarkblue, align=center, minimum width=2.25cm, minimum height=0.9cm, font=\scriptsize},
    llm/.style={rectangle, rounded corners, draw=AIRIblue, fill=AIRIpaper, text=AIRIdarkblue, align=center, minimum width=2.0cm, minimum height=0.9cm, font=\scriptsize},
    fusion/.style={rectangle, rounded corners, very thick, draw=AIRIgreen, fill=AIRIpaper, text=AIRIdarkblue, align=center, minimum width=2.0cm, minimum height=0.9cm, font=\scriptsize},
    output/.style={rectangle, rounded corners, very thick, draw=AIRIgreen, fill=AIRIpaper, text=AIRIdarkblue, align=center, minimum width=2.0cm, minimum height=0.9cm, font=\scriptsize},
    supervisedarrow/.style={-{Latex[length=2.5mm]}, thick, draw=AIRIgreen},
    llmarrow/.style={-{Latex[length=2.5mm]}, thick, draw=AIRIblue}
]
\node[font=\small\bfseries, text=AIRIgreen] at (6.0,4.15) {Supervised candidate generation};
\node[font=\small\bfseries, text=AIRIgreen, anchor=west] at (12.35,2.5) {Selective LLM invocation};
\node[font=\small\bfseries, text=AIRIgreen, anchor=west] at (3.7,0.1) {Fusion and final prediction};

\node[input] (textnew) at (0.0,3.4) {Clinical diagnosis\\text};
\node[supervised, minimum width=2.8cm] (encodernew) at (3.2,3.4) {Text preprocessing\\+ ModernBERT encoder};
\node[supervised, minimum width=2.1cm] (letternew) at (6.0,3.4) {Leading-letter\\classifier (RF)};
\node[supervised, minimum width=2.4cm] (parentnew) at (8.9,3.4) {Conditional parent\\classifier (HGB)};
\node[candidates] (candnew) at (11.9,3.4) {Supervised top-10\\candidates};

\node[nocall] (rankernew) at (8.2,1.65) {Candidate ranker\\(no LLM call)};
\node[llm] (gatenew) at (11.9,1.65) {Pre-LLM\\gate};
\node[llm] (reranknew) at (14.8,1.65) {LLM candidate\\selector};
\node[fusion] (fusionnew) at (11.9,0.1) {Pre-LLM\\fusion};
\node[output] (outnew) at (14.8,0.1) {Final parent\\code};

\draw[supervisedarrow] (textnew.east) -- (encodernew.west);
\draw[supervisedarrow] (encodernew.east) -- (letternew.west);
\draw[supervisedarrow] (letternew.east) -- (parentnew.west);
\draw[supervisedarrow] (parentnew.east) -- (candnew.west);

\coordinate (candidatefork) at (11.9,2.65);
\draw[thick, draw=AIRIgreen] (candnew.south) -- (candidatefork);
\draw[supervisedarrow] (candidatefork) -- (8.2,2.65) -- (rankernew.north);
\draw[llmarrow] (candidatefork) -- (gatenew.north);
\draw[llmarrow] (gatenew.east) -- (reranknew.west);

\draw[supervisedarrow] (rankernew.south) |- (fusionnew.west);
\draw[llmarrow] (reranknew.south) -- (14.8,0.85) -- (11.9,0.85) -- (fusionnew.north);
\draw[supervisedarrow] (fusionnew.east) -- (outnew.west);
\end{tikzpicture}
\caption{Final call-efficient pipeline for ICD-10 parent-code prediction. The supervised candidate generator produces a shared top-10 list. A no-call candidate-ranker route and a selectively invoked LLM-reranker route then converge in the Pre-LLM fusion selector, which outputs the final parent code. Green denotes supervised or no-call processing, while blue denotes selectively invoked LLM processing.}
\label{fig:pipeline}
\end{figure*}
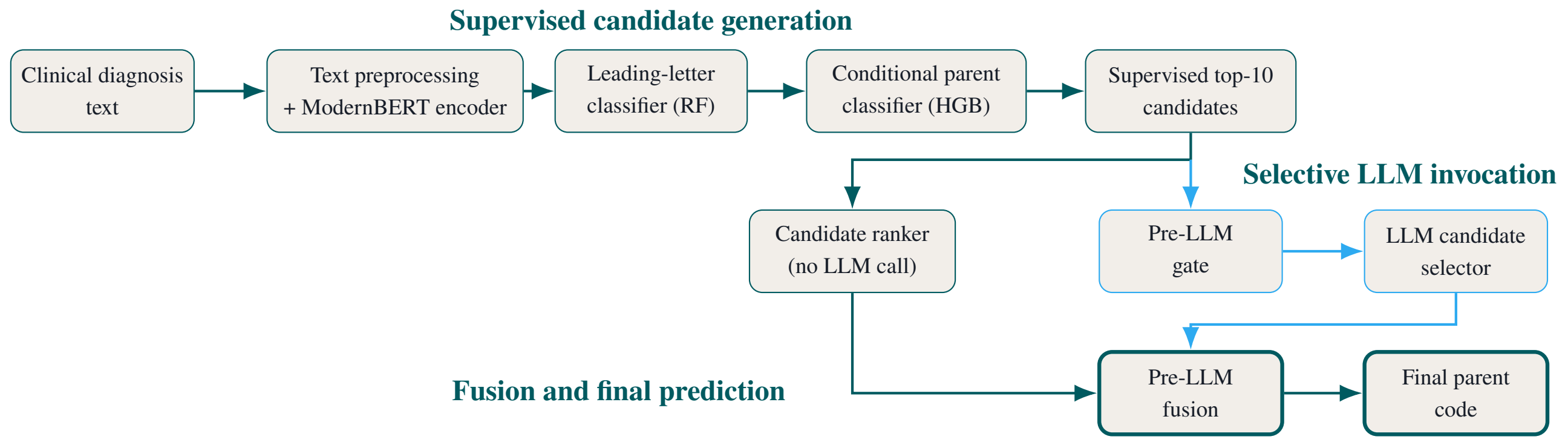

In the final selective setting, the same supervised top-10 candidate list is also used by three deployable selectors: Pre-LLM gate, Candidate ranker, and Pre-LLM fusion. The gate decides whether an LLM call is worthwhile, the candidate ranker provides a no-call reranking alternative, and the fusion selector combines these two routes into the final deployable prediction. The structure of the final prediction pipeline is shown in Figure~\ref{fig:pipeline}.

\subsection{Embedding extraction}

The proposed method constructs a two-level hierarchical embedding space for ICD-10 codes. The first level produces a fixed-length representation (92 dimensions) useful for letter prediction, while the second level produces a letter-specific representation tailored for rubric prediction. Both levels are built purely from pairwise BERT-based similarity scores without requiring external labelled data for training the embedding itself.

We begin with a set of \(N = 9930\) ICD-10 diagnosis-code pairs in Russian from the Electronic reference book~\cite{mkb10su}. These codes are first partitioned into 23 groups according to their leading letter (the first character of the code). Letters P, U, V, W, X, Y in the test set, and U, V, W, X in the training set had fewer than 4 samples. We therefore excluded them and did not compute embeddings for these letters. Within each letter group, codes are further subdivided into rubrics -- defined as the part of the code before the decimal point (e.g., ``A01'' in ``A01.2'').

For each rubric within a given letter group, we compute pairwise semantic similarity scores between all diagnosis descriptions using the distilled ModernBERT matching model described above. This yields a similarity matrix for each rubric. For each diagnosis \(d_i\) in the rubric, we calculate its mean similarity to all other diagnoses in the same rubric:

\[
\bar{s}_i = \frac{1}{m-1} \sum_{j \neq i} s(d_i, d_j),
\]

where \(m\) is the number of diagnoses in the rubric. For rubric containing a single dictionary description (\(m=1\)), that description is used directly as its centroid. We then select the diagnosis with the highest \(\bar{s}_i\) as the second-level centroid for that rubric. This process is repeated for every rubric in each letter group, producing a set of second-level centroids for each rubric.

Next, we aggregate these second-level centroids by their parent letter group. Within each letter group, we again compute pairwise similarities among all second-level centroids, and for each centroid we compute its average similarity to the others. We sort the centroids by this average and retain the top-4 centroids as the first-level centroids for that letter. Across all 23 letters, this yields a total of \(23 \times 4 = 92\) first-level centroids.

Given a new diagnosis either from test and train datasets described as a text string \(x\):

1. First-level embedding:  
   We compute the similarity between \(x\) and each of the 92 first-level centroids \(c^{(1)}_k\):

   \[
   e^{(1)}_k = s(x, c^{(1)}_k), \quad k = 1, \dots, 92.
   \]

   The resulting 92-dimensional vector
   \[
   \mathbf{e}^{(1)} = [e^{(1)}_1, \dots, e^{(1)}_{92}]
   \]
   is the first-level embedding. This vector is used to train a Random Forest classifier that predicts the leading letter of the ICD-10 code.

2. Second-level embedding:
   Once the predicted letter \(L\) is known (either from the classifier or from the ground truth), we retrieve the set of second-level centroids associated with that letter. Let these be \(\{c^{(2)}_1, \dots, c^{(2)}_{n_L}\}\), where \(n_L\) is the number of rubrics (and thus centroids) for letter \(L\).  
   We compute similarities to each of these centroids:

   \[
   e^{(2)}_j = s(x, c^{(2)}_j), \quad j = 1, \dots, n_L.
   \]

   The vector \(\mathbf{e}^{(2)} = [e^{(2)}_1, \dots, e^{(2)}_{n_L}]\) is the second-level embedding. Because \(n_L\) varies across letters, we train a separate classifier for each letter to predict the rubric (the part before the decimal point) from this embedding.

\subsection{Selective reranking and deployable selectors}
To compare the supervised pipeline with alternative formulations from the literature, we also evaluate retrieval and LLM reranking baselines. The final reranking experiments use Qwen3.5-122B-A10B, as recorded in the saved validation outputs. The retrieval baseline ranks category candidates within the ICD-10 letter group using the available candidate scores and returns top-k categories. Diagnosis text and ICD-10 category descriptions are encoded with ModernBERT, and candidates are ranked by cosine similarity between the diagnosis embedding and the precomputed category embeddings. The LLM reranking baseline then passes the diagnosis and the top-10 candidates to Qwen3.5-122B-A10B, asking the model to choose the best code from the candidate list. This evaluates whether a large language model can improve candidate selection when the search space is restricted.

A second LLM experiment uses the supervised conditional category stage rather than the retrieval baseline as the candidate generator. In this setting, the supervised model provides top-10 category candidates and Qwen3.5-122B-A10B reranks them into a top-5 list. This experiment tests whether LLM reranking can improve an already strong supervised candidate list. For the LLM reranking experiments, the prompt asks the model to select a single best code from the provided top-10 candidate list. To report Top-k accuracy and MRR@5, we derive an induced ranking by placing the LLM-selected candidate at rank one and the remaining candidates in their original order. If the LLM returns a code outside the candidate list or an unparseable response, the sample is treated as having no valid prediction: Top-k accuracy is 0 for all k, and the reciprocal rank is 0 for MRR@5.

We also evaluate GPT-5.6-Sol as a direct zero-shot coding baseline. Unlike the candidate-constrained rerankers, this model receives the diagnosis text and is asked to predict the ICD-10 diagnosis directly, without using the supervised candidate list. This baseline therefore invokes a large, comparatively expensive model for every test example and has a 100\% LLM call rate.

In addition to these baselines, the final deployable version of the system uses three selective components built on top of the supervised top-10 candidate list. \textit{Pre-LLM gate} predicts whether an LLM call is needed for a given diagnosis and therefore controls inference cost. \textit{Candidate ranker} is a no-call reranker that reorders the supervised candidates using only supervised and lexical features. \textit{Pre-LLM fusion} combines the outputs of the gate-driven LLM path and the candidate ranker to produce the final deployable prediction under the same cost budget as the gate.

\subsection{Evaluation protocol}
We report letter-level accuracy, F1 (micro and weighted), per-class precision and recall, category Top-1/Top-3/Top-5 accuracy, MRR@5, rare-versus-frequent metrics, analysis of the transitions of the predicted and real letters, and calibration diagnostics. The rare-class analysis uses the support-based split (letters with support more than 250 are considered as frequent, other represented letters are considered as rare). For LLM-based experiments, we additionally report valid-candidate rate, because the model may sometimes return a code outside the provided candidate list.

For the final selective methods, we additionally report locked-test Top-1 accuracy, OOF-CV Top-1 accuracy, and LLM call rate, because deployment cost is a primary constraint. Model and threshold selection are performed with out-of-fold cross-validation, while the locked test set is used only for final reporting.

For metrics computed from fixed model predictions, uncertainty is estimated with a nonparametric bootstrap over test-set rows. Each bootstrap iteration resamples test objects with replacement and recomputes the metric from the already saved predictions; models are not retrained. Unless stated otherwise, error bars in metric plots correspond to 95\% bootstrap confidence intervals.

\section{Results}
\subsection{Letter-level diagnostics}
\begin{figure}[t]
  \centering
  \widefig{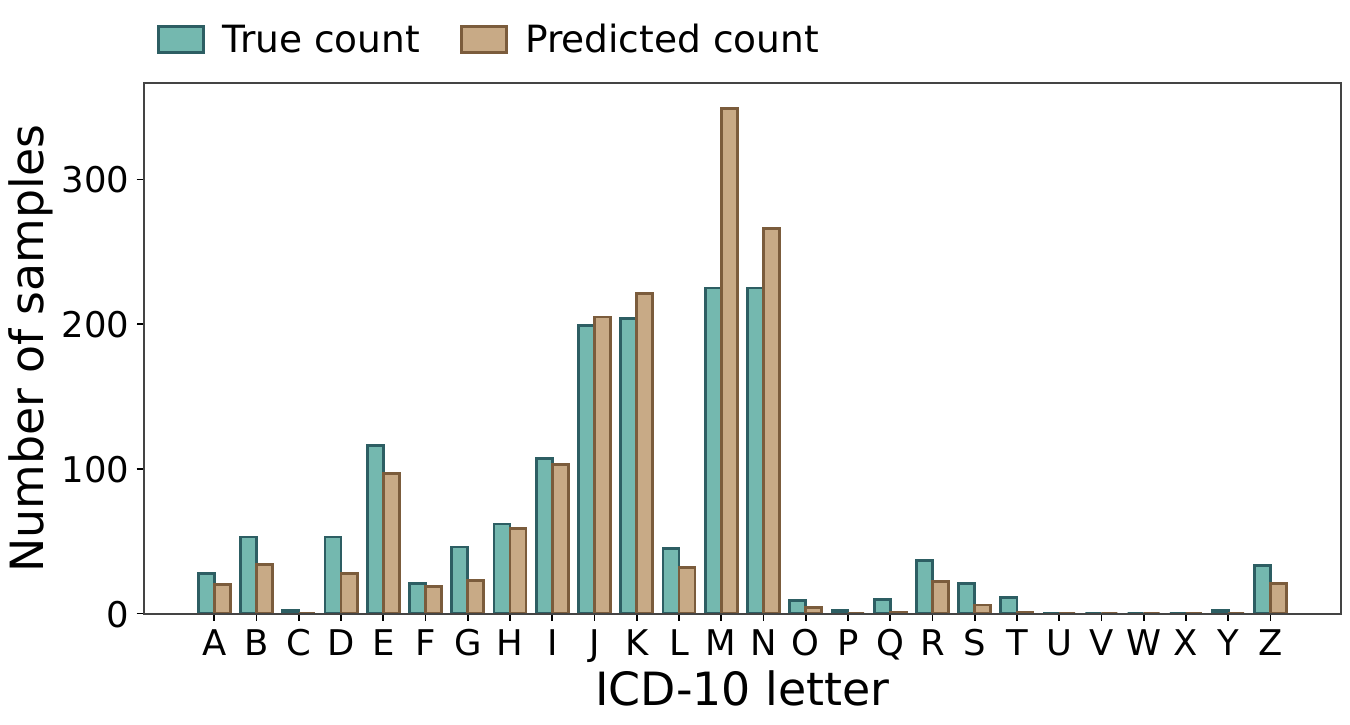}
  \par\smallskip
  \textbf{(a)} True and predicted ICD-10 letter counts.
  \par\medskip
  \widefig{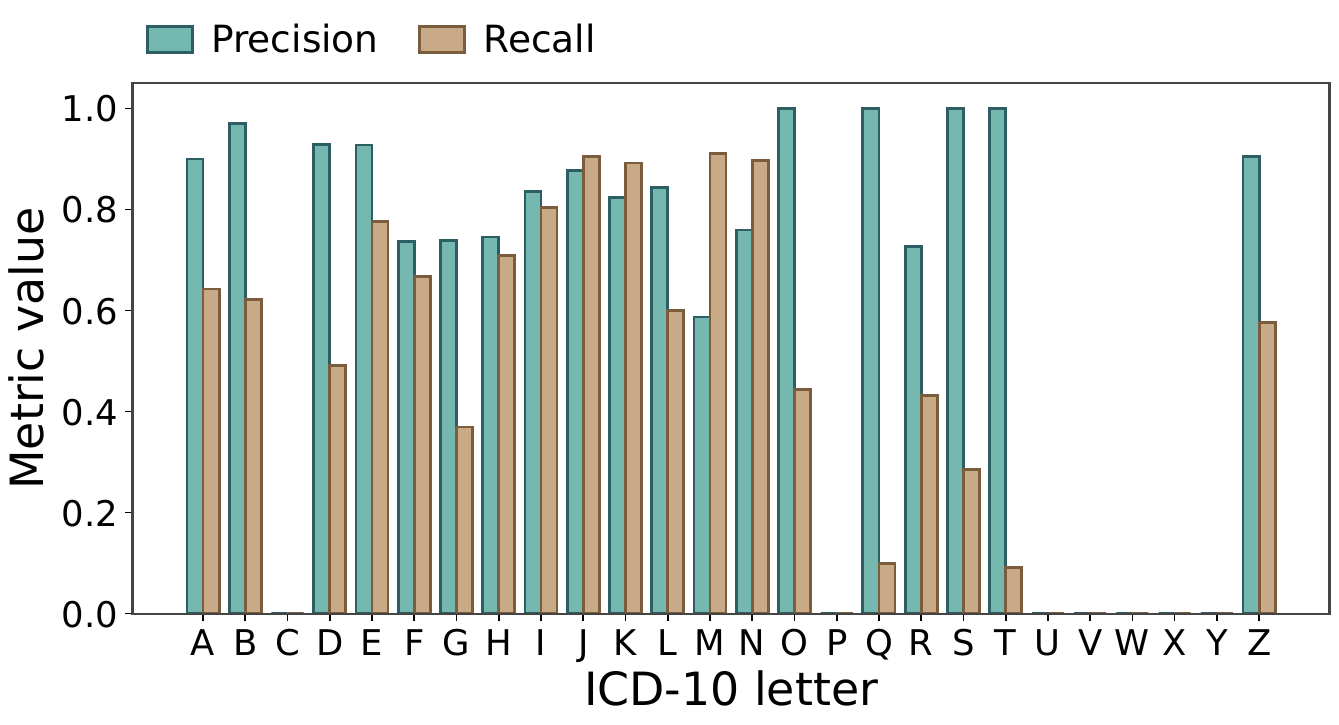}
  \par\smallskip
  \textbf{(b)} Per-letter precision and recall.
  \caption{Letter-level distribution and per-class quality diagnostics. Frequent letters are overrepresented among predictions, while rare letters are underrepresented.}
  \label{fig:letter_distribution_quality}
\end{figure}

The letter classifier reaches 77.50\% Top-1 and 85.57\% Top-3 accuracy on the locked test set. The main downstream target is parent-code prediction, but letter-level errors remain important because the parent classifier is conditional on the predicted broad group. The diagnostic plots in Figure~\ref{fig:letter_distribution_quality} show that frequent letters are more stable, while rare letters remain harder to predict.

\begin{figure}[t]
  \centering
  \widefig{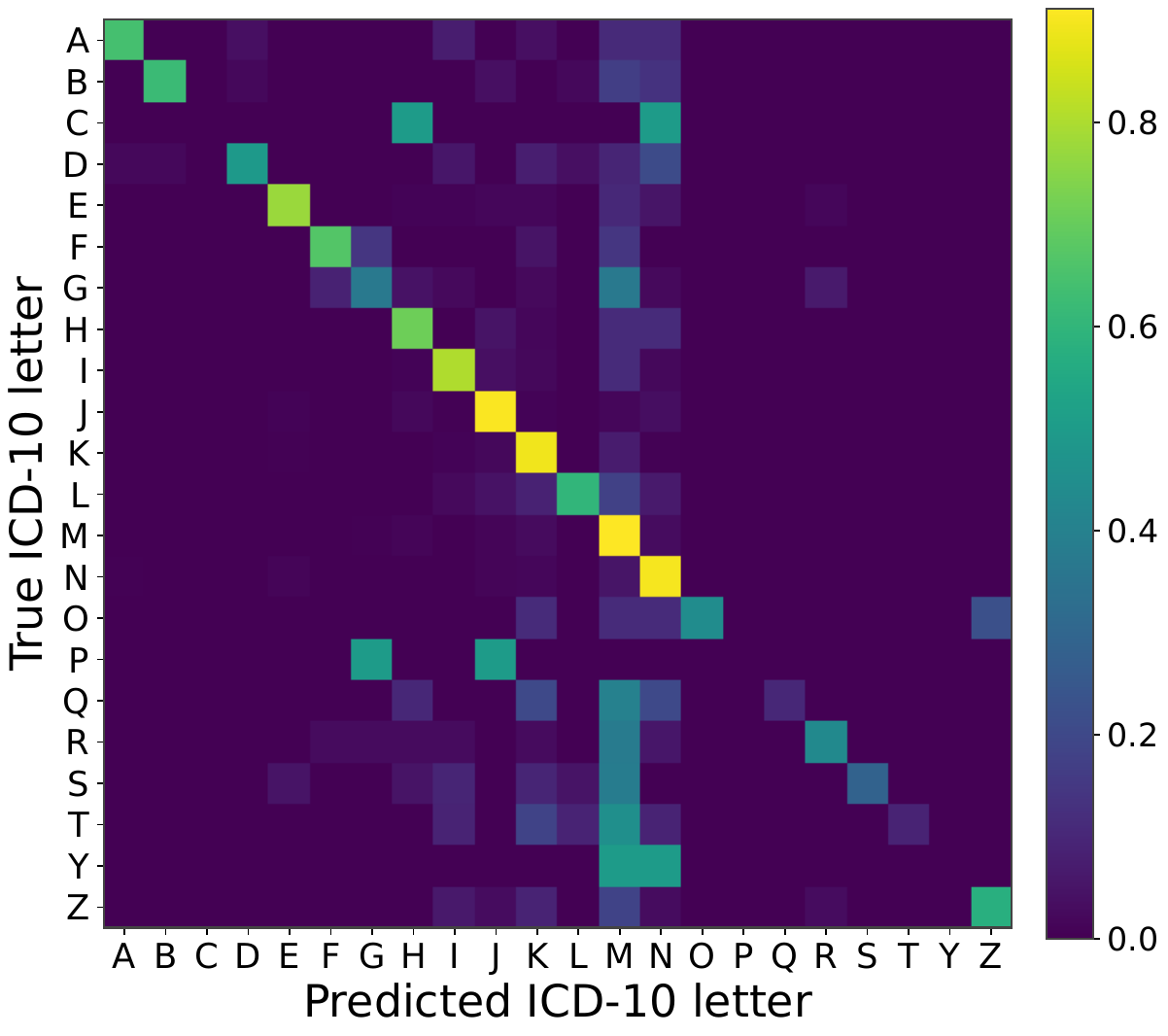}
  \caption{Normalized confusion matrix for ICD-10 letter prediction. Rows correspond to true letters and columns correspond to predicted letters. Off-diagonal cells reveal systematic error destinations.}
  \label{fig:confusion_matrix}
\end{figure}

The normalized confusion matrix in Figure~\ref{fig:confusion_matrix} further shows where letter-level errors are concentrated. Its off-diagonal cells identify systematic confusion routes rather than isolated mistakes.

\begin{figure}[htbp]
  \centering
  \midwidefig{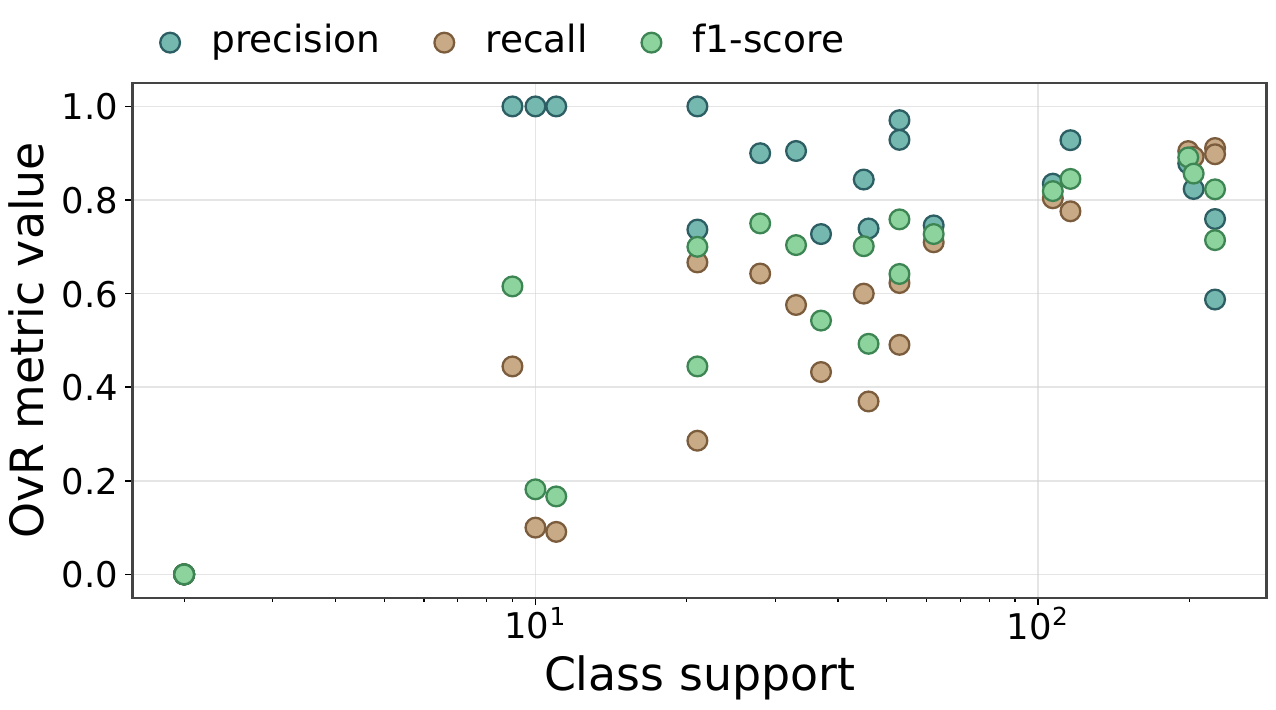}
  \caption{Relationship between class support and OvR quality metrics. Recall and F1-score increase strongly with support, while precision is less support-dependent.}
  \label{fig:support_vs_ovr}
\end{figure}

The support-performance relationship in Figure~\ref{fig:support_vs_ovr} confirms that rare letters remain the main reliability issue. Recall and F1-score increase with class support, whereas precision is less support-dependent.

\subsection{Parent-code candidate quality}
Table~\ref{tab:parent_candidate_results} reports the parent-code candidate quality before final selection. The supervised conditional parent stage reaches 74.32\% Top-1 accuracy on the locked test, but its Top-10 coverage is 93.05\%. Calling the LLM for every row improves Top-1 accuracy to 80.41\%, Top-5 accuracy to 92.46\%, and MRR@5 to 85.85\%, while Top-10 coverage remains 93.05\% because the reranker only reorders the supervised candidate set. This setting requires one LLM call per diagnosis.

\begin{table*}[t]
\centering
\normalsize
\begin{tabular}{@{}lccccc@{}}
\toprule
Method & Top-1 & Top-3 & Top-5 & Top-10 & MRR@5 \\
\midrule
Supervised parent stage & 74.32\% & 88.15\% & 91.00\% & 93.05\% & 81.42\% \\
Full LLM selection & 80.41\% & 91.46\% & 92.46\% & 93.05\% & 85.85\% \\
\bottomrule
\end{tabular}
\caption{Parent-code candidate-generation and reranking quality on the locked test set. The full-invocation LLM setting serves as an accuracy reference and requires an LLM call for every case; it is not the final selective system.}
\label{tab:parent_candidate_results}
\end{table*}

\subsection{Final deployable methods}
The final comparison is shown in Table~\ref{tab:final_results}. OOF-CV is used for model and threshold selection, while the locked test is used only for final reporting. The direct zero-shot GPT-5.6-Sol baseline reaches 81.01\% locked-test accuracy but requires a large-model call for every diagnosis. Pre-LLM fusion is the strongest deployable method: it reaches 81.24\% OOF-CV accuracy and 83.19\% locked-test accuracy while preserving the Pre-LLM gate's 31.97\% LLM call rate. It therefore exceeds the zero-shot GPT-5.6-Sol result by 2.18 percentage points while using 68.03\% fewer LLM calls. The candidate ranker is weaker than Pre-LLM fusion, but it is useful as a no-call alternative because it improves over the supervised parent-stage Top-1 baseline without any LLM requests.

\begin{table}[!ht]
\centering
\small
\begin{tabular}{@{}lccc@{}}
\toprule
Method & OOF-CV & Locked test & LLM calls \\
\midrule
Zero-shot GPT-5.6-Sol & --- & 81.01\% & 100.00\% \\
\midrule
Pre-LLM gate & 79.51\% & 82.13\% & 31.97\% \\
Candidate ranker & 77.42\% & 78.56\% & 0.00\% \\
Pre-LLM fusion & \textbf{81.24\%} & \textbf{83.19\%} & 31.97\% \\
\bottomrule
\end{tabular}
\caption{Final parent-code prediction results, including the direct zero-shot large-model reference. OOF-CV is used for model and threshold selection; locked-test accuracy is the final held-out score. LLM calls are reported on the locked test set; an em dash denotes that OOF-CV was not evaluated for the zero-shot reference.}
\label{tab:final_results}
\end{table}

\begin{figure}[htbp]
  \centering
  \midwidefig{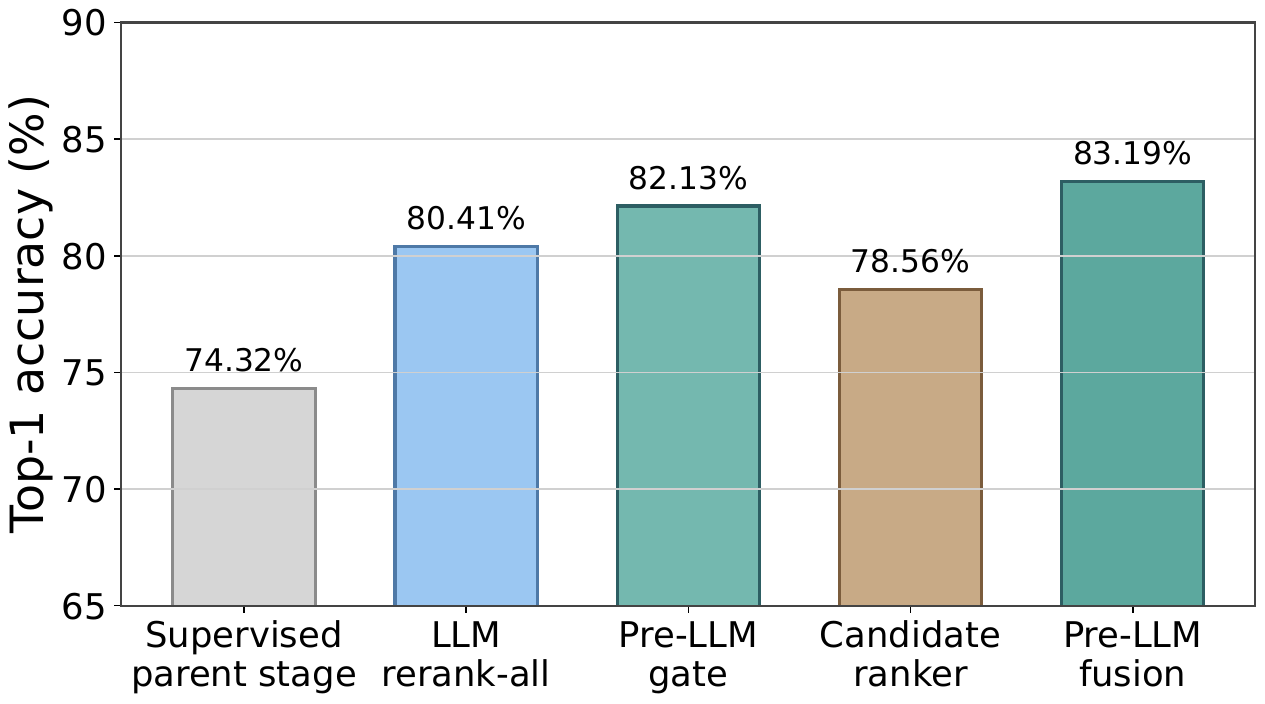}
  \caption{Locked-test Top-1 accuracy across supervised, full-LLM, and deployable selective variants. Pre-LLM fusion provides the strongest deployable result.}
  \label{fig:locked_comparison}
\end{figure}

Figure~\ref{fig:locked_comparison} places the final methods in context. Pre-LLM fusion improves locked-test Top-1 accuracy from 74.32\% for the supervised parent stage to 83.19\%, reducing the error rate from 25.68\% to 16.81\%.

\subsection{Rare-class and calibration diagnostics}
\begin{figure}[!ht]
  \centering
  \includegraphics[width=0.92\linewidth]{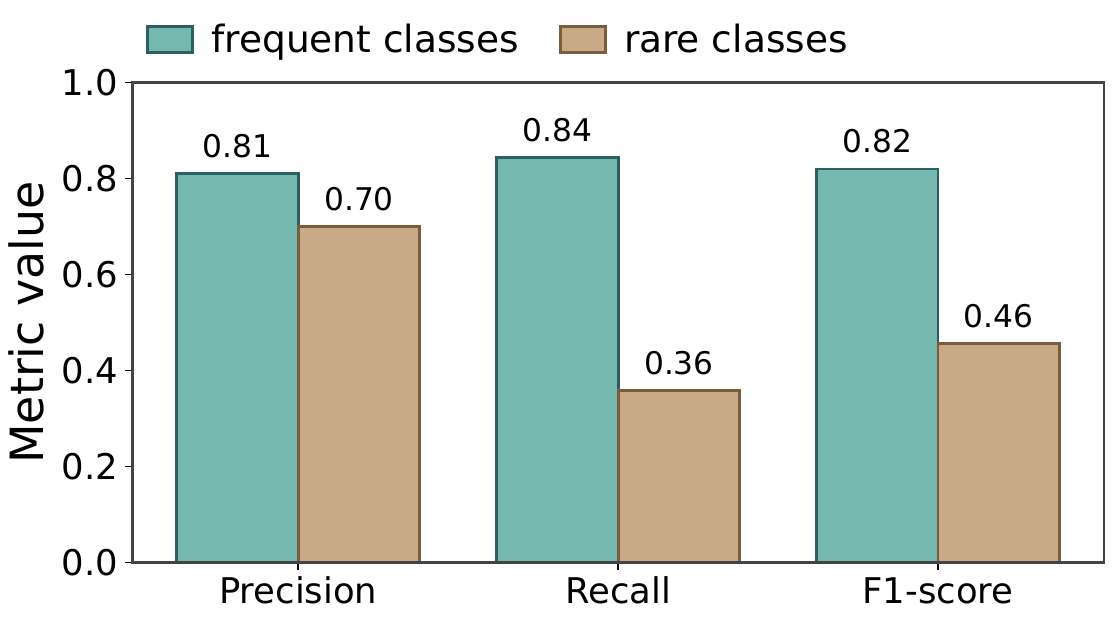}
  \par\smallskip
  \textbf{(a)} Rare-versus-frequent ICD-10 letter metrics.
  \par\medskip
  \includegraphics[width=0.92\linewidth]{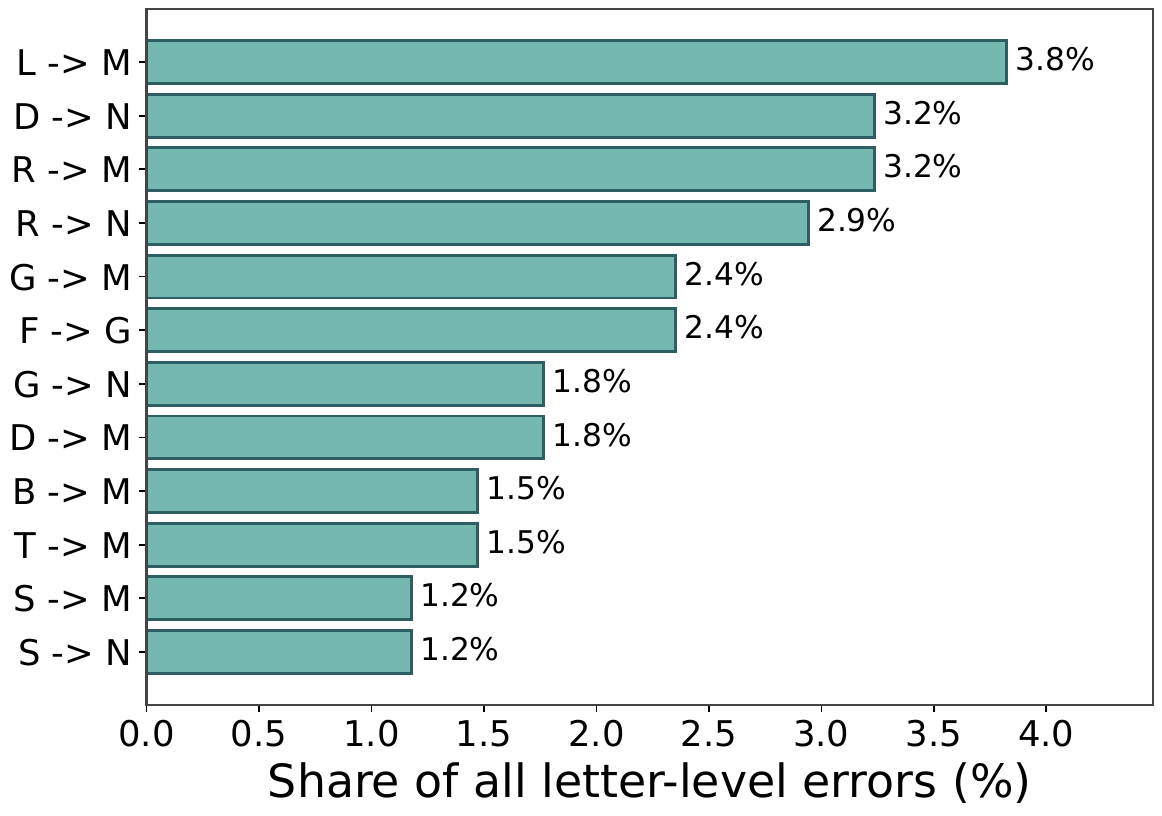}
  \par\smallskip
  \textbf{(b)} Most frequent letter-level error transitions.
  \caption{Rare-class behavior and dominant error routes. Rare classes are most strongly affected through reduced recall and F1-score, and their errors are often redirected to frequent ICD-10 letters.}
  \label{fig:rare_error_diagnostics}
\end{figure}

The final accuracy gain does not remove the underlying rare-class problem. Rare classes still have lower recall and lower F1-score, and many rare-letter errors are redirected to frequent letters. This behavior is summarized in Figure~\ref{fig:rare_error_diagnostics}. These diagnostics are important because aggregate accuracy can improve even when low-support classes remain unreliable.

\begin{figure}[!ht]
  \centering
  \makebox[\linewidth][c]{\includegraphics[width=0.78\linewidth]{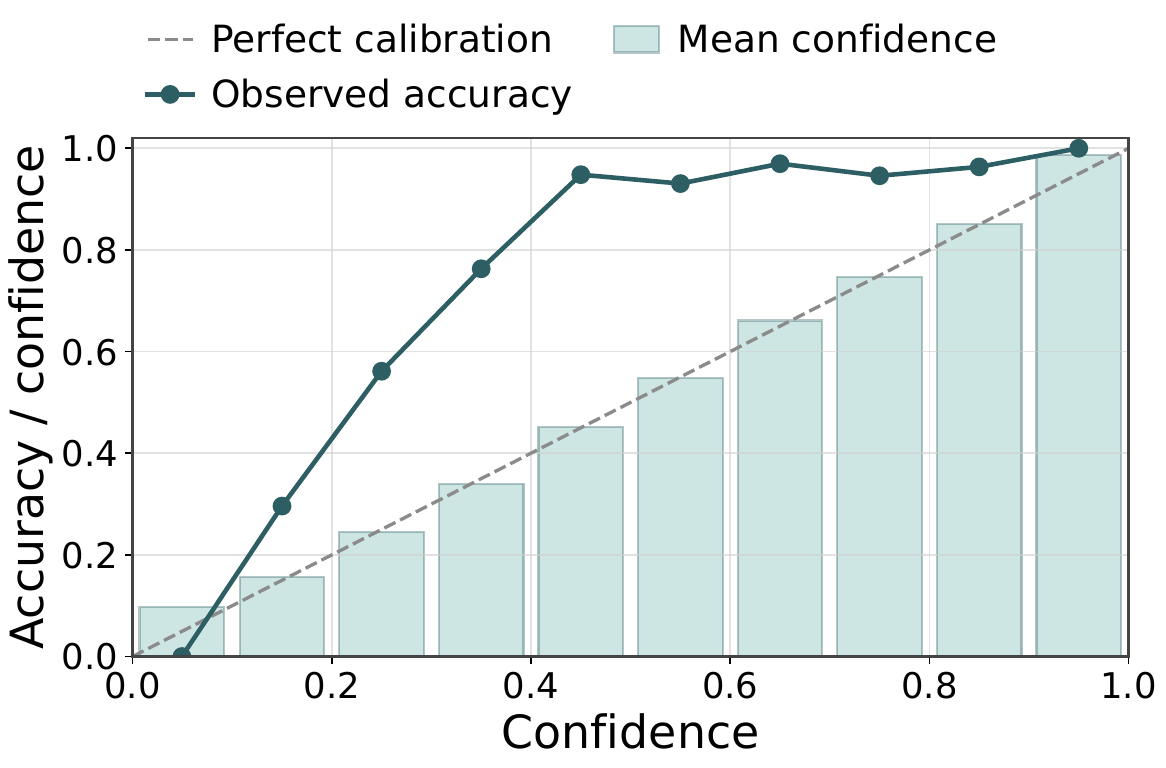}}
  \caption{Reliability diagram for letter-level predictions. Confidence is informative, but the model is not fully calibrated across all probability bins.}
  \label{fig:reliability}
\end{figure}

The reliability diagram in Figure~\ref{fig:reliability} remains useful for deployment planning. Model confidence separates correct from incorrect predictions on average, but is not fully calibrated across all probability bins. Therefore, raw confidence should not be interpreted as a calibrated probability without additional validation.

\section{Discussion}
The experiments show three main findings. First, the supervised parent-code candidate generator is a strong and necessary base model. Its Top-10 coverage of 93.05\% means that the correct parent code is usually present in the candidate list, even when the top-1 prediction is wrong. This high candidate coverage is what makes downstream reranking effective.

Second, the LLM is useful but should not be called blindly. Full LLM reranking improves locked-test Top-1 accuracy from 74.32\% to 80.41\%, but it requires 100\% LLM calls. Similarly, direct zero-shot GPT-5.6-Sol coding reaches 81.01\% while invoking this large, comparatively expensive model for every diagnosis. The Pre-LLM gate learns to identify cases where an LLM request is likely to help and reaches 82.13\% accuracy with 31.97\% LLM calls. This is both more accurate and more cost-efficient than either full-invocation setting in the final validation evaluation.

Third, the best deployable method is Pre-LLM fusion. It combines the selective LLM gate with the no-call candidate ranker and reaches 83.19\% locked-test accuracy. The OOF-CV score of 81.24\% is close enough to the locked-test result to support the validation protocol: model selection is performed on cross-fitted predictions, while the locked test remains a held-out evaluation set.

The candidate ranker is important even though it is not the best standalone method. Its locked-test accuracy is 78.56\% with zero LLM calls, which makes it a useful fallback and a useful input to fusion. It also shows that candidate-level features contain signal beyond the original supervised top-1 ranking.

\section{Limitations}
The study is based on the currently available project dataset and should not be interpreted as a direct state-of-the-art comparison against RuCCoD or MIMIC-based ICD coding systems. The datasets, languages, label spaces, and input formats differ. The purpose of the external comparison is to position the task and baselines, not to claim superiority over systems evaluated on different corpora.

The final pipeline also depends on the specific candidate generator and LLM reranking prompt. A different candidate generator, prompt, or hosted LLM could change the balance between the gate, the ranker, and the fusion selector. The present results should therefore be interpreted as a validated pipeline result for this dataset and this candidate-reranking setup, rather than as a universal property of all LLM-assisted ICD coding systems.

Finally, the rare-class analysis identifies the main remaining failure mode but does not yet solve it. Future work should compare rare-class-aware training strategies, including class weighting, resampling, hierarchical losses, calibrated abstention, and explicit open-set handling for unseen parent codes.

\section{Conclusion}
This study presents a final call-efficient ICD-10 parent-code prediction pipeline based on supervised candidate generation, selective LLM reranking, a candidate ranker, and a fusion selector. The final fusion system reaches 81.24\% OOF-CV accuracy and 83.19\% locked-test accuracy while invoking the LLM for only 31.97\% of cases. It also exceeds the 81.01\% accuracy of direct zero-shot GPT-5.6-Sol coding, which requires a large-model call for every diagnosis. Compared with the supervised parent-code baseline, the locked-test error rate decreases from 25.68\% to 16.81\%.

The main practical conclusion is that LLMs are most useful as controlled rerankers inside a strong supervised candidate-generation pipeline. The supervised model provides high candidate coverage, the pre-LLM gate limits online invocation, the candidate ranker supplies a no-call alternative, and the fusion selector combines them into the strongest deployable result.

\bibliography{aaai2027_paper}

\end{document}